\documentclass[a4paper]{jpconf}
\usepackage{amsthm, amssymb, amsmath,bbm}
\usepackage{color}
\usepackage{myMacros}
\begin{document}
\title{Interrelation of equivariant Gaussian processes and convolutional neural networks}

\author{Andrey Demichev and Alexander Kryukov}

\address{Skobeltsyn Institute of Nuclear Physics, M.V. Lomonosov Moscow State University,\\
	1(2) Leninskie gory, Moscow 119991, Russian Federation}

\ead{demichev@theory.sinp.msu.ru, kryukov@theory.sinp.msu.ru}



\begin{abstract}

Currently there exists rather promising new trend in machine leaning (ML) based on the relationship between neural networks (NN) and  Gaussian processes (GP), including many related subtopics, e.g., signal propagation in NNs, theoretical derivation of learning curve for NNs, QFT methods in ML, etc. An important feature of convolutional neural networks (CNN) is their equivariance (consistency) with respect to the symmetry transformations of the input data. In this work we establish a relationship between the many-channel limit for CNNs equivariant with respect to two-dimensional Euclidean group with vector-valued neuron activations and the corresponding independently introduced equivariant Gaussian processes (GP).
 
\end{abstract}

\section{Introduction}

In recent years, a correspondence has been established between the appropriate asymptotics of deep neural networks (DNNs), including convolutional ones (CNNs), and the machine learning methods based on Gaussian processes (GP), see \cite{novak2018bayesian} and refs. therein. The ultimate goal of establishing such interrelations is to achieve a better theoretical understanding of various methods of machine learning (ML) and their improvement.  Since Gaussian processes are mathematically similar to Euclidean quantum field theory (QFT), one of the intriguing consequences of this correspondence is the potential for using a vast arsenal of QFT methods to analyze deep neural networks \cite{cohen2019learning}, in particular for predictions of learning curves and DNN outputs.

An important feature of convolutional networks is their equivariance (consistency) with respect to the symmetry transformations of the input data \cite{cohen2016steerable,kondor2018clebsch}. Equivariance guarantees that exactly the same filters are applied to each part of the input image regardless of position and that the network can detect any given object equally well regardless of its location respecting data symmetry properties. 

It is important that the above mentioned works on establishing the interrelations between CNNs and GPs deal only with translational equivariance of images. On the other hand there exists investigations of more general (e.g., including rotations) equivariant neural Gaussian processes \cite{holderrieth2020equivariant} but without established relations with CNNs in the appropriate limit. In our preceding work \cite{demichev2021pos}, initiated in the framework of Russian--German Astroparticle Data Life Cycle Initiative \cite{BDD}, we derived the many-channel limit for a toy 1D CNNs with $SO(2)$ symmetry and scalar activations keeping explicit equivariance at each step of the derivation and calculated the corresponding equivariant GP kernel. In the present work we provide the derivation of the many-channel limit for the practically important case of CNNs equivariant with respect to Euclidean group of motions of a two-dimensional input space and vector-valued neuron activations. Such DNNs are described in the framework of steerable convolutional neural networks (SCNN) \cite{cohen2016steerable, weiler2019general} which use induced representations of the respective symmetry groups. The result obtained can be applied for theoretical analysis of CNNs and improvement of their performance  with the account of a prior information about symmetries of input data (see, \textit{e.g.}, \cite{cohen2019learning} for such an analysis without account of any symmetry).

\section{Many-channel CNNs with vector-valued activations and $SE(2)$ symmetry \label{sec:MCC}}


We consider a series of $L+1$ convolutional layers, $l = 0, \dots, L$ in a steerable convolutional neural network (SCNN) \cite{cohen2016steerable}  with $SE(2)\equiv T_2\rtimes SO(2)$ symmetry on a $2D$-plane \cite{weiler2019general}. Here $T_2$ and $SO(2)$ are groups of two-dimensional translations and rotations, respectively. Such a SCNN defines feature spaces as spaces of  fields $y:\R^2\to\R^c$ which associate a $c$-dimensional feature vector $y({\vec r})\in\R^c$ to each point $\vec r$ of a base space, in the considered case the plane $\R^2$. The transformation law of a general feature field $f:\R^2\to\R^c$ is fully characterized by the $SE(2)$ group representation $\rho:\ SO(2)\mapsto GL(\R^c)$ specifying how the $c$ channels of each feature vector $y(\vec r)$ mix under transformations. A simplest example are scalar feature fields ${s:\R^2\to\R}$ with the trivial representation $\rho(g) = \1$. Such feature fields describe, e.g., gray-scale images, temperature, pressure, etc. Infinite-channel limit for CNNs with scalar fields was derived in the works \cite{novak2018bayesian}, and the issue of equivariance (for 1D analog) was discussed in \cite{demichev2021pos}. Vector fields ${\By:\R^2\to\R^2}$ transform as ${\By(\vec r)\mapsto g\cdot \By\left(g^{-1}(\vec r-\vec t)\right)}$, where $g$ is the matrix of two-dimensional rotations. Examples include fields of wind direction and speed, electric fields, etc. More details and examples can be found, e.g., in \cite{weiler2019general} (in the context of SCNNs) and in \cite{holderrieth2020equivariant} (in the context of GPs). In this paper we consider the case of vector feature fields.

More generally, fields transforms under the induced representation (see \cite{weiler2019general} and refs. therein)
\begin{align}\label{eq:induced_rep}
y^l(\vec r)\ \ \mapsto\ \ 
\left(\left[\operatorname{Ind}_{SO(2)}^{T_2\rtimes SO(2)}\rho\right]\!\!(tg)\cdot y^l\right)\! (\vec r)
\ \ :=\ \ \rho(g)\cdot y^l\left(g^{-1}(\vec r-\vec t)\right).
\end{align}
It is important that the $\rho$-representation can be reducible. Moreover, as it is well known, all complex-valued irreducible representattion (irreps) of the Abelian group $SO(2)$ are one-dimensional.

The most general \emph{equivariant linear map} between steerable feature spaces (SCNN's layers), transforming under $\rho_\text{in}$ and $\rho_\text{out}$, is given by convolutions with $SO(2)$-steerable kernels (filters)
$\omega:\R^2\to\R^{c_\text{out}\times c_\text{in}}$, satisfying a kernel constraint (see \cite{worrall2017harmonic},  \cite{weiler2019general} and refs therein)
\begin{align}\label{eq:kernel_constraint}
\omega(g\vec r)\ =\ \rho_\text{out}(g)\omega(\vec r)\rho_\text{in}(g^{-1}) \quad\forall g\in SO(2),\ \vec r\in {\R}^2 \,.
\end{align}

From the point of view of limiting transition to Gaussian processes, SCNNs poses the following main problems to be solved: (1)
	the application-important vectors (like, e.g., wind speed and direction) are also treated as channels, so the question is how one can go to the \textit{infinite}-channel limit given that such vectors must have a \textit{finite} dimensionality defined by an application task;
	(2) since the filter $\omega(\vec r)$ satisfies the condition (\ref{eq:kernel_constraint}), its values are related at different points of the base space $\R^2$ and cannot be initialized independently, the latter being an essential ingredient for the derivation of the infinite-channel limit.

The first problem is solved in the present work by using in (\ref{eq:induced_rep}) a \textit{reducible} representation so that real-valued activations have two indices $y^l_{i\alpha}(\vec r)$, the first one (latin) numerating $n^l$ different irreps within the $\rho$-representation, while the second (Greek) index indicating components within the $i$-th two-dimensional irrep. In the group theory the number of irreps $n^l$ is often called multiplicity. Thus the infinite-channel limit can be harmlessly carried out over the multiplicity $n^l$ keeping the application-meaningful vector dimensionality fixed. The second problem is solved by separation of longitudinal and angle modes of the filter $\omega(\vec{r})$ and intializing independently only the longitudinal modes which are not constraint by the condition (\ref{eq:kernel_constraint}) (see below).

An essential technical simplification of the infinite-channel derivation can be achieved by using instead of two-dimensional vectors $y_{i\alpha}(\vec{r})\ (\alpha=1,2;\, i=1,\dots,n^l)$ complex-valued feature field  $y_{i}(\vec{r}):\ \R^2\mapsto\C$. In particular, this can be clearly seen from a comparison of the solution derivation for the constraint (\ref{eq:kernel_constraint}) in the complex \cite{worrall2017harmonic} and real-valued \cite{weiler2019general} forms. Certainly, in general two-dimensional vectors and matrices are not fully equivalent to complex numbers. In particular, a general $2\times 2$ matrix $\omega_{ij,\alpha\beta}$ (for fixed $i,j$) cannot be represented as a complex number. But the filter $\omega$ satisfying (\ref{eq:kernel_constraint}) proves to be a unique angular frequency (mode) \cite{weiler2019general}
\begin{align}\label{eq:filter_mode}
\omega_{ij}(\vec r)\ = R_{ij}(r)\left[ 
\begin{array}{cc}
\cos((m-n)\phi) & -\sin((m-n)\phi) \\
\sin((m-n)\phi) & \cos((m-n)\phi)
\end{array}  \right]\ ,
\end{align}
where $\{r,\phi\}$ are the polar coordinates for the vector $\vec{r}$. This matrix and its action on a two-dimensional vectors can be represented as the multiplication of two complex numbers: $\sum_j \omega_{ij}(\vec{r})y_j$; $\omega_{ij},y_j\in\C$. Further discussion on the relation between real and complex valued representations in the context of SCNNs can be found in \cite{weiler2019general}, Appendix~F.5.

The network has activations $y^{l}(\vec{r})$ and pre-activations $z^{l}(\vec{r})$ for each input $x_i(\vec{r})$. They are defined by the following mappings from layer to layer:
\begin{align} \label{eq:def_relations_RS}
y^{l}_{i}(\vec{r}) &= \left\{\begin{array}{cc} x_{i}(\vec{r}) &  l=0 \\
\varphi^{(CS)}\left( z^{l-1}_{i}(\vec{r}) \right) &  l > 0 \end{array}\right.,
&
z^{l}_{i}(\vec{r}) &= \sum_{j=1}^{n^{l}} [\omega_{ij}\star y_j](\vec{r}) \ ,
\end{align}
the bar over a character denotes the complex conjugate and $\varphi^{(CS)}$ is a nonlinearity in the coordinate space. As it is seen, the pre-activations $z^{l}_{i}(\vec{r})$ are defined by the cross-correlation
$\star$ on $\R^2$: $[\omega_{ij}\star y_j](\vec{r}) \equiv \int d^2r^\prime\,\bar{\omega}_{ij}(\vec{r}^{\,\prime})y_j(\vec{r}+\vec{r}^{\,\prime})$. One can show that the cross-correlation in (\ref{eq:def_relations_RS}) with a filter satisfying (\ref{eq:kernel_constraint}) preserve the equivariance of the SCNN.

However since we have to detach angular modes, working with the mapping (\ref{eq:def_relations_RS}) in coordinate space proves to be very cumbersome, in particular because we have to sum vectors in polar coordinates in the cross-correlation, \textit{cf.} \cite{baddour2009operational}. To overcome this difficulty, we use Fourier transform of the feature fields 
$F(\vec{p})=\int d^2r\,\exp\left\{-\i\, \vec{r}\vec{p}\right\}f(\vec{r})$ (here $\{f,F\}$ stands for $\{y,Y\}$, $\{z,Z\}$ or $\{\omega,\Omega\}$, respectively).  In the terms of the Fourier transform the induced representation (\ref{eq:induced_rep}) reads 
\begin{align} \label{eq:induced_rep_FS}
Y^{l}_{i}(\vec{p}) &\mapsto  Y^{\prime\,l}_{i}(\vec{p}) = \erom^{\i\,(k\theta-\vec{t}\vec{p})}\, Y^{l}_{i}(g^{-1}_\theta\vec{p})\ ,
\end{align}
where $\vec{t}$ is the parameter of the translation subgroup $T_2\in SE(2)$, $\theta$ is the parameter of the $SO(2)$ subgroup and $k$ is the index of the induced representation\footnote{It is worth noting that while the coordinate space representations (\ref{eq:induced_rep}) are defined in the space of functions on $\R^2=SE(2)/SO(2)$, the Fourier space representations (\ref{eq:induced_rep_FS}) are defined in the space of functions on a circle $S^1=SE(2)/T_2$. Details on the latter representation can be found in \cite{vilenkin1978special}.}, \textit{cf.} (\ref{eq:induced_rep}).

Since we have to detach the angular modes of the filter $\omega,\Omega$, we expand it and the feature fields over the angles: $F(\vec{p})=\sum_mF_m(p)\,\exp\left\{ -\i\,m\psi\right\}$, where $\{p,\psi\}$ are the polar coordinates for $\vec{p}$. In these Fourier-space and angular-mode fields the mappings (\ref{eq:def_relations_RS}) reads as follows
\begin{align} \label{eq:def_relations_FS}
Y^{l}_{i,n}(p) &= \left\{\begin{array}{ccc} X_{i,n}(p) &  & l=0 \\
\varphi^{(FS)}\left( Z^{l-1}_{i,\cdot}(p) \right) &  & l > 0 \end{array}\right.,
&
Z^{l}_{i,n}(p) &= \sum_{j=1}^{n^{l}} {\OMM}_{ij,q_{l}}(p)Y^l_{j,n+q_l}(p) \ .
\end{align}
The $\varphi^{(FS)}$ denotes a nonlinearity in the Fourier space, $q_l$ is an arbitrarily chosen mode of the filter $\Omega$ so that the latter  satisfies the constraint (\ref{eq:kernel_constraint}), precisely because of the selection of a unique mode. It is worth noting that the angular modes in the coordinate and Fourier spaces for the same mode numbers are related to each other by the Hankel transform \cite{baddour2009operational}. Transformation properties of the angular modes can be easily deduced from (\ref{eq:induced_rep_FS}) but they are rather cumbersome and we do not display them here. As it is readily seen, the cross-product in (\ref{eq:def_relations_RS}) is substituted in (\ref{eq:def_relations_FS}) by the the pointwise product; this is a consequence of the well-known convolution theorem.

For the illustration of the general approach we choose the \textit{polynomial} nonlinearity in the Fourier space 
preserving the explicit equivariance of the SCNN (\textit{cf.} \cite{kondor2018clebsch} and \cite{weiler2019general}):
$Y^l_{i}(\vec{p})= \varphi^{(FS)}\left( Z^{l-1}_{i}(p) \right) \equiv \bar{Z}^{l-1}_i(\vec{p})Z^{l-1}_i(\vec{p})Z^{l-1}_i(\vec{p})\ ,$ 
or, equivalently, in terms of the angular modes
\begin{align} \label{eq:nonlinearity_angular}
Y^l_{i,m}(p)&= \varphi_m^{(FS)}\left( Z^{l-1}_{i,\cdot}(p) \right) \equiv\sum_{n,k}\,\bar{Z}^{l-1}_{i,k}(p)Z^{l-1}_{i,n}(p)Z^{l-1}_{i,m+k-n}(p)\ .
\end{align}
Actually, the latter can be presented as a 1D discrete cross-product $\star_1$ (\textit{cf.} \cite{baddour2009operational}) with respect to angular mode indices: $Y^l_{i,m}(p) = [Z^{l-1}_{i,\cdot}(p)\star_1Z^{l-1}_{i,\cdot}(p)\star_1Z^{l-1}_{i,\cdot}(p)]$.

According to the general approach to NN-GP correspondence \cite{novak2018bayesian}, it is assumed the Gaussian prior on the \textit{independent} filter weights in the Fourier space, 
$\Omega^{l}_{ij,q_l}(p) \sim \cN\left(0, {\sigma^2_\omega}/{n^{l}}\right)\ ,  $
where $\sigma^2_\omega$ is the weight variance. 
Since $\Omega^l_{ij,q_l}(p)$ are \textit{i.i.d.} variables, they satisfy
$\E[\Omega^l_{ij,q_l}(p)\Omega^l_{i^\prime j^\prime,q_l}(p^\prime)] = \E[\bar{\Omega}^l_{ij,q_l}(p)\bar{\Omega}^l_{i^\prime j^\prime,q_l}(p^\prime)]=0$; 
$\E[\bar{\Omega}^l_{ij,q_l}(p)\Omega^l_{i^\prime j^\prime,q_l}(p^\prime)] = \E[\Omega^l_{ij,q_l}(p)\bar{\Omega}^l_{i^\prime j^\prime,q_l}(p^\prime)]=({\sigma^2_w}/{2n^l})\delta_{i i^\prime}\delta_{j j^\prime}\delta_{pp^\prime}.$
The pre-activations $Z^l$ are linear combinations of the Gaussian variables $\Omega^l$, specified by the previous layer's activations $Y^{l}$. Thus for the conditioned on $Y^{l}$ pre-activations one has
$\E[Z_{i,n^\prime}^l(p^\prime)\bar{Z}_{i,n}^l(p)]=\frac{\sigma^2_w}{2}\delta_{pp^\prime}K^l_{n+q_l,n^\prime+q_l}(p,p^\prime)$,
where the uncentered covariance matrix $K^{l}$ of the activations $Y^l$ is defined as (\textit{cf.} \cite{novak2018bayesian}):
\begin{align}\label{eq:covMatrix}
K^{l}_{n,n'}\left(p, p'\right) &\deff \frac 1 {n^l}\sum_{i=1}^{n^l} Y^{l}_{i,n}(p)\bar{Y}^{l}_{i,n'}(p') \ .
\end{align}
It is important to emphasize that in general $K^{l}$ is constructed as an outer product of two two-dimensional vectors $K^{l}\sim \bY\otimes\bY$ and cannot be represented as a complex valued function. But we are looking for an \textit{equivariant} Gauss kernel which is presumably can be obtained as an appropriate limit for $K^{l}$. A key observation, greatly simplifying the following calculations, is that  according to Theorem~1 of the paper \cite{holderrieth2020equivariant} an equivariant Gauss kernel satisfies the same constraint condition (\ref{eq:kernel_constraint}) as the SCNN's filter. Therefore we suppose that the covariance matrix $K^{l}$ is of the form (\ref{eq:filter_mode}) and it can indeed be represented as a complex  valued function.

Since a linear combination of Gaussian variables is itself a Gaussian, we can conclude that 
$\left( Z^l | Y^{l} \right) \sim \cN_\cC(0, \widetilde{\Gamma})$, 
where $\cN_\cC(0, \widetilde{\Gamma})$ is the circularly-symmetric central complex normal distribution, and the covariance $\widetilde{\Gamma}$ reads
$\widetilde{\Gamma}^l_{n,n'}(p,p^\prime) =\frac{\sigma^2_w}{2}\delta_{pp^\prime}K^l_{n+q_l,n^\prime+q_l}(p,p^\prime)\otimes I_{n^{l+1}}$.
Thus the normal distribution of $\left(Z^l | Y^{l} \right)$ depends only on $K^l$ and similarly to \cite{novak2018bayesian} we can conclude that the random variable $\left(Z^l | K^l\right)$ has the same distribution: $(Z^l | K^{l}) \sim \cN_\cC(0, \widetilde{\Gamma})$.

Now by means of the relation (\ref{eq:def_relations_FS}) we can express the covariance (\ref{eq:covMatrix}) via the pre-activations, take $n^l\rightarrow\infty$ limit and use the weak law of large numbers (\textit{cf}. \cite{novak2018bayesian}):
\begin{align}\label{eq:n-limit}
& K^l_{n,n^\prime}(p,p^\prime) = \frac{1}{n^l}\sum_{i=1}^{n^l}{\varphi}_n^{(FS)}\left(\bar{Z}^{l-1}_{i,\cdot}(P),Z^{l-1}_{i,\cdot}(p)\right) \bar{\varphi}_{n^\prime}^{(FS)}\left(\bar{Z}^{l-1}_{i,\cdot}(p^\prime),Z^{l-1}_{i,\cdot}(p^\prime)\right) \nbr
& \limlar{n^l}{\infty} \E_{Z\sim \cN_{\cC}(0,\Gamma^{l-1})} \left[{\varphi}_n^{(FS)}\left(\bar{Z}^{l-1}_{i,\cdot}(P),Z^{l-1}_{i,\cdot}(p)\right) \bar{\varphi}_{n^\prime}^{(FS)}\left(\bar{Z}^{l-1}_{i,\cdot}(p^\prime),Z^{l-1}_{i,\cdot}(p^\prime)\right)\right] 
\ , 
\end{align}
where 
$\Gamma^{l-1}_{n,n'}(p,p^\prime) =({\sigma^2_w}/{2})\delta_{pp^\prime}K^l_{n+q_l,n^\prime+q_l}(p,p^\prime)$,
and we have used the nonlinearity (\ref{eq:nonlinearity_angular}). The mean value in (\ref{eq:n-limit}) is equal to the Gaussian-like integral which can be calculated by adding the terms with external sources and using Feynman technique. The result allows us to show that: 
\begin{itemize}
	\item if $K^{l-1}_{n,n'}(p,p')=\delta_{nn'}\delta_{pp'}K^{l-1}_n(p)$ (diagonal kernel), then the same is true for $K^l$; this property guarantees the translational invariance for the kernel, \textit{cf.} \cite{holderrieth2020equivariant}, Theorem 1;
	\item if $K^{l-1}(\vec{p})$ contains only \textit{one} angular mode, say with some number $s_{l-1}$, 
	then the same is true for $K^l$; this property guarantees the rotational equivariance (\cite{holderrieth2020equivariant}, Theorem 1), similarly to the solution for the filter with the constraint (\ref{eq:kernel_constraint}).
\end{itemize}
As a result of the calculations, for the $K^{l}_n(p)$ the simple recursive relation has been found
\begin{align}\label{eq:recursiveRel}
K_{s}^L(p)=\left(6\left({\sigma_w^2}/{2}\right)^3\right)^{(3^L-1)/2}\Big(K^0_{s+q_0+\dots +q_L}(p)\Big)^{3^L}\ .
\end{align}
Thus we have derived the self-consistent solution for the equivariant kernel of the Gaussian process obtained as $n_l\rightarrow\infty$ limit of the SCNN. The development of a general method for such a derivation was the main goal of this work.

\section{Conclusion\label{sec:Con}}
In this work we have derived the many-channel limit for $2D$ steerable CNNs with $SE(2)\equiv T_2\rtimes SO(2)$ symmetry and vector-valued neuron activations with explicit equivariance at each step of the derivation and calculated the corresponding equivariant GP kernel. All the subtleties and mathematically rigorous proofs for the expressions obtained are quite similar to the case of classical CNNs and can be found in \cite{novak2018bayesian}. Thereby this work filled the gap between many-channel equivariant CNNs and independently introduced equivariant GPs \cite{holderrieth2020equivariant}. 

In order to derive the limit for application-important vector-valued feature fields we separated channels indices in two categories: the indices that numerate the vector components within an irrep and used to describe their transformations under matrix representations of a symmetry group and the indices that numerate different irreducible representations (of the same or different types). The second type of the indices are not restricted and can be used for the limiting transition to the corresponding GP. Also it was technically important to use the Fourier space for convenient detachment of independent components of SCNN's filters. 

In the subsequent works the result obtained will be applied for theoretical analysis of equivariant CNNs with the aim of improvement of their performance.

\ack{This work was funded by the Russian Science Foundation (grant No. 22-21-00442).}

\medskip
\section*{References}
\bibliographystyle{iopart-num}
\bibliography{Demichev_ACAT21}

\end{document}